\documentclass[lettersize,journal]{IEEEtran}
\usepackage{amsmath,amsfonts}
\usepackage{algorithmic}
\usepackage{algorithm}
\usepackage{array}
\usepackage[caption=false,font=normalsize,labelfont=sf,textfont=sf]{subfig}
\usepackage{textcomp}
\usepackage{stfloats}
\usepackage{url}
\usepackage{verbatim}
\usepackage{graphicx}
\usepackage{cite}
\usepackage{multirow}
\usepackage{amsfonts,amssymb} 
\usepackage{enumerate}
\usepackage{color}
\usepackage[dvipsnames, svgnames, x11names]{xcolor}
\usepackage[normalem]{ulem}
\useunder{\uline}{\ul}{}
\usepackage{pifont}
\DeclareSymbolFont{mathbf}{OT1}{cmr}{bx}{n}
\DeclareMathSymbol{X}{\mathalpha}{mathbf}{`X}
\DeclareMathSymbol{Z}{\mathalpha}{mathbf}{`Z}
\DeclareMathSymbol{Y}{\mathalpha}{mathbf}{`Y}
\DeclareMathSymbol{W}{\mathalpha}{mathbf}{`W}
\DeclareMathSymbol{E}{\mathalpha}{mathbf}{`E}
\begin{document}

\title{M$^2$CD: A Unified MultiModal Framework for Optical-SAR Change Detection with Mixture of Experts and Self-Distillation}

\author{Ziyuan Liu, Jiawei Zhang, Wenyu Wang, and Yuantao Gu,~\IEEEmembership{Senior Member,~IEEE}
\thanks{Ziyuan Liu, Jiawei Zhang and Yuantao Gu are with the Department of Electronic Engineering, Beijing National Research Center for Information Science and Technology, Tsinghua University, Beijing 100084, China (e-mail: liuziyua22@mails.tsinghua.edu.cn; jiawei-z23@mails.tsinghua.edu.cn; gyt@tsinghua.edu.cn). Wenyu Wang is with the College of Communications Engineering, Army Engineering University of PLA, Nanjing 210007, China (e-mail: wwy@aeu.edu.cn). (\textit{Corresponding author: Yuantao Gu.})}

}



\maketitle
\begin{abstract}  
Most existing change detection (CD) methods focus on optical images captured at different times, and deep learning (DL) has achieved remarkable success in this domain. However, in extreme scenarios such as disaster response, synthetic aperture radar (SAR), with its active imaging capability, is more suitable for providing post-event data. This introduces new challenges for CD methods, as existing weight-sharing Siamese networks struggle to effectively learn the cross-modal data distribution between optical and SAR images. 
To address this challenge, we propose a unified MultiModal CD framework, M$^2$CD. 
We integrate Mixture of Experts (MoE) modules into the backbone to explicitly handle diverse modalities, thereby enhancing the model's ability to learn multimodal data distributions.
Additionally, we innovatively propose an Optical-to-SAR guided path (O2SP) and implement self-distillation during training to reduce the feature space discrepancy between different modalities, further alleviating the model's learning burden. 
We design multiple variants of M$^2$CD based on both CNN and Transformer backbones. Extensive experiments validate the effectiveness of the proposed framework, with the MiT-b1 version of M$^2$CD outperforming all state-of-the-art (SOTA) methods in optical-SAR CD tasks.  
\end{abstract}


\section{INTRODUCTION}
Change detection (CD) in remote sensing (RS) images is a critical technology used to detect and analyze surface changes. It plays a significant role in various fields, such as land use monitoring, urban development, and natural disaster assessment.  

In recent years, with the rise of deep learning (DL), architectures such as convolutional neural networks (CNNs)\cite{fcsn,stanet,ifn,snunet,bit,tinycd,hanet,cgnet}, transformers\cite{changeformer,changer}, and foundation models\cite{ttp} have achieved remarkable success in the field of CD. Despite their diverse structures and training strategies, most of these methods follow the paradigm of Siamese neural networks. 
Specifically, they employ two weight-shared sub-networks to process the pre-event and post-event images separately. The features extracted from these two branches are then fused by a detector to identify change regions. 
This weight-sharing dense architecture effectively reduces the number of model parameters.

Optical RS images are the most commonly used modality in CD tasks. However, optical sensors rely on passive sensing technology, which often faces challenges under unfavorable imaging conditions, such as during flood or wildfire events. 
In contrast, synthetic aperture radar (SAR) is an active microwave sensor capable of all-weather and all-day ground observation\cite{dsrkd}, making it more suitable for providing post-event images under extreme conditions for CD. 
This requires handling data with significant modality differences from different sensors, posing substantial challenges to existing CD models.

Li et al. \cite{sm3det} points out that when handling cross-modal data, the learning capacity of existing dense network structures is limited, as their shared weights struggle to adapt to the distributions of different modalities. 
This issue is also prevalent in multimodal optical-SAR CD. 
To address this challenge and provide algorithmic support for rapid disaster response using SAR images, we propose a unified \textbf{M}ulti\textbf{M}odal \textbf{CD} framework (M$^2$CD). 
By introducing modality-specialized Mixture of Experts (MoE)\cite{moe} modules into the backbone and innovatively proposing an Optical-to-SAR transition path (O2SP) for self-distillation guidance, we reduce the feature space discrepancies between different modalities and alleviate the model's burden in processing multimodal data.
Our main contributions are as follows:  
\begin{enumerate}
    \item The proposed M$^2$CD framework is highly versatile and robust, compatible with various backbone architectures such as CNNs and Transformers.
    \item The introduced MoE and O2SP significantly enhance the model's capability to process Optical-SAR data. In particular, O2SP is only utilized during training, avoiding additional computational overhead during inference, thus offering a novel paradigm for multimodal CD.
    \item Extensive experiments on the CAU-Flood dataset demonstrate that M$^2$CD outperforms all state-of-the-art (SOTA) methods. Ablation studies further validate the effectiveness of both MoE and O2SP in improving model performance.
\end{enumerate}

\begin{figure*}[!t]
    \centering
    \includegraphics[width=7in]{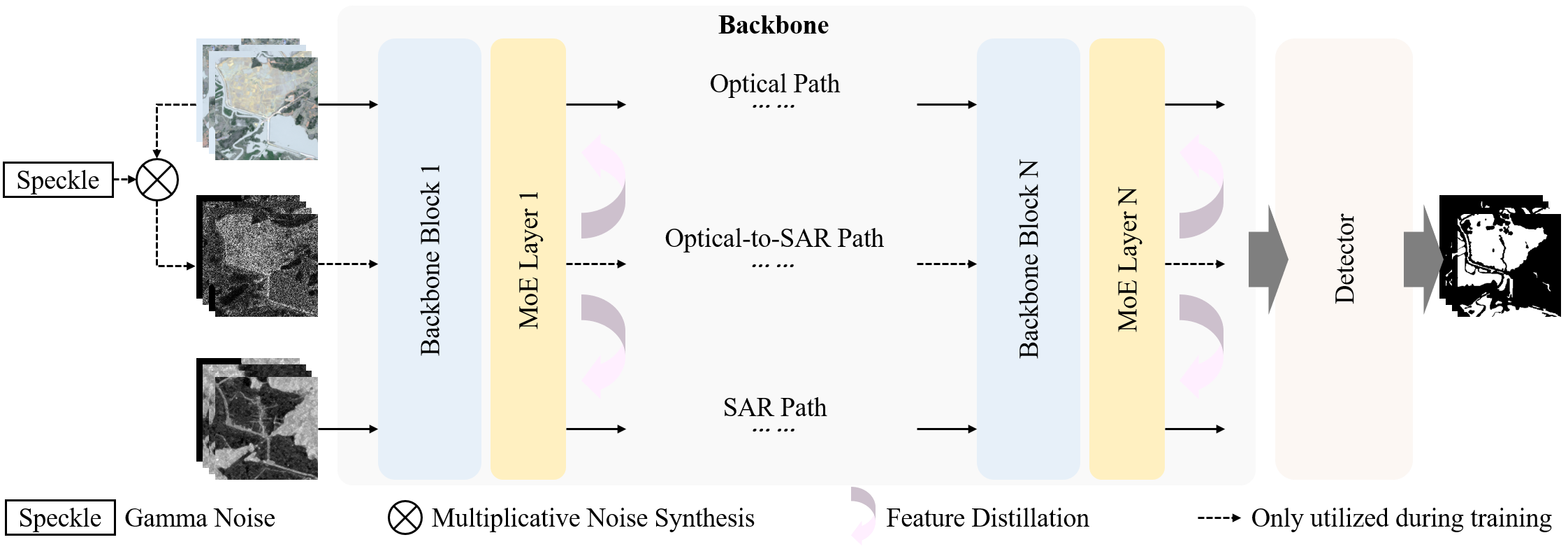}
    \caption{Overview of the proposed M$^2$CD framework.}
    \label{fig:architecture}
\vspace{-0.3cm}
\end{figure*}

\section{Proposed Method} \label{A}

\subsection{Overview of the M$^2$CD Framework}  
Mainstream DL-based CD methods typically rely on a dual-branch Siamese architecture. 
Specifically, the pre-event image $X_1$ and the post-event image $X_2$ are fed into a weight-shared backbone $\mathcal{B}$, generating a series of features $\{Z_1^{(1)}, \dots, Z_1^{(N)}\} = \mathcal{B}(X_1)$ and $\{Z_2^{(1)}, \dots, Z_2^{(N)}\} = \mathcal{B}(X_2)$. 
These features are then fused by a detector $\mathcal{D}$ to predict the change label $\hat{Y} = \mathcal{D}(\{Z_1^{(1)}, \dots, Z_1^{(N)}\}, \{Z_2^{(1)}, \dots, Z_2^{(N)}\})$. 
However, in optical-SAR CD, the significant modality differences caused by different sensors severely degrade model performance. 
To address this issue, we propose the M$^2$CD framework, as illustrated in Fig. \ref{fig:architecture}, which introduces the following improvements to the existing CD paradigm:  

\textbf{MoE:} We integrate MoE into the backbone to explicitly process distinct modalities. Multiple experts adaptively handle images from different temporal phases, effectively alleviating the backbone's burden in processing cross-modal data.  

\textbf{O2SP:} Unlike the classic dual-branch structure, we innovatively introduce the Optical-to-SAR Path (O2SP) in addition to the Optical Path (OP) for the pre-event image and the SAR Path (SP) for the post-event image. By incorporating self-distillation, O2SP bridges the feature representations of the two modalities, further narrowing the gap between optical and SAR data.

\subsection{MoE Module}
For the intermediate feature $Z_t^{(i)}$ output by the $i$-th layer of the backbone, the corresponding MoE layer operates as follows:  
\begin{align}
\tilde{Z}_t^{(i)} \!&=\! \sum_{k=1}^M G(Z_t^{(i)}) \cdot \Phi_k(Z_t^{(i)}),\nonumber
\\
G(Z_t^{(i)}) \!&=\! \text{Top}_k\!\left(\!\text{Softmax}\!\left(\frac{E^T W Z_t^{(i)}}{\|W Z_t^{(i)}\|_2 \left\|E\right\|_2}\!\right)\!\right)\!, t\!\in\!\{1,\!2\},\nonumber
\end{align}
where $M$ denotes the total number of experts, $G$ represents the gating function, and the expert $\Phi$ is implemented as a $1 \times 1$ convolution in our framework. The learnable embedding matrix is denoted as $E$. Given an input feature $Z_t^{(i)}$, it is first transformed by the weight matrix $W$. The transformed feature $W Z_t^{(i)}$ is then compared with each expert embedding in $E$ to compute similarity scores, which are normalized by the product of the norms of $W Z_t^{(i)}$ and $E$ to ensure scale invariance. These similarity scores are subsequently processed by a Softmax function to generate a probability distribution, representing the relevance of each expert to the input feature. Finally, the $\text{Top}_k$ operator selects the $k$ most relevant experts based on the highest probabilities, reweighting their outputs while setting the remaining experts to zero.

This design creates a sparser feature space within the backbone model, enabling distinct representation learning for optical and SAR modalities. By preventing feature space interference and effectively leveraging cross-modal knowledge, the MoE module significantly enhances CD performance.

\vspace{-0.3cm}
\subsection{O2SP Self-Distillation} 

Unlike optical RS images, SAR images are often severely contaminated by speckle noise due to the coherent summation of many elementary echoes within the radar resolution cell, which represents one of the most significant distinctions between the two modalities. Based on the fully developed speckle assumption, SAR images can be modeled as $\mathbf{I} = \mathbf{R} \odot \mathbf{S}$, where $\mathbf{I}$ represents the real-world SAR image, $\mathbf{R}$ denotes the clean SAR image (which is not available in reality), $\mathbf{S}$ represents the multiplicative speckle noise, and $\odot$ denotes the element-wise multiplication. Each element $s$ of the speckle noise $\mathbf{S}$ is typically assumed to follow a Gamma distribution with a mean of 1 and a variance inversely proportional to the number of looks $L$: $p(s) = L^L s^{L-1} e^{-L s} / \Gamma(L)$,
where $\Gamma$ denotes the Gamma function.  
As illustrated in Fig. \ref{fig:architecture}, we leverage the fully developed speckle assumption to generate simulated SAR images from the pre-event optical images. These simulated SAR images serve as an intermediate representation between optical RS images and real-world SAR images, acting as a bridge to connect the two distinct modalities. 
We employ O2SP to guide the distillation of features from the OP and SP in the feature space, thereby reducing the gap between the modalities. 
After passing through the backbone blocks and MoE modules, OP and SP obtain multi-level features $\{\tilde{Z}_1^{(1)}, \dots, \tilde{Z}_1^{(N)}\}$ and $\{\tilde{Z}_2^{(1)}, \dots, \tilde{Z}_2^{(N)}\}$, respectively. Similarly, the features obtained from O2SP are denoted as $\{\tilde{Z}_{\text{tran}}^{(1)}, \dots, \tilde{Z}_{\text{tran}}^{(N)}\}$. 
The self-distillation loss is introduced to minimize the $L_1$ distance between OP, O2SP, and SP:  
\begin{equation}
\mathcal{L}_{\text{SD}} = \sum_{i=1}^N \left( \left\|\tilde{Z}_{\text{tran}}^{(i)} - \tilde{Z}_1^{(i)}\right\|_1 + \left\|\tilde{Z}_{\text{tran}}^{(i)} - \tilde{Z}_2^{(i)}\right\|_1 \right).
\end{equation}

In addition to the self-distillation loss, we employ a cross-entropy (CE) loss function to supervise the entire model training using ground truth (GT) labels:  
\begin{equation}
\label{ce_loss}
\mathcal{L}_{\text{CE}} =  -y \log(\hat{y}) - (1 - y) \log(1 - \hat{y}),
\end{equation}
where $y$ represents the GT label and $\hat y$ denotes the predicted label. The CE loss~\eqref{ce_loss} is computed pixel-wise and averaged over the whole image.
The complete loss function for training is defined as:  
\begin{equation}
\mathcal{L} = \mathcal{L}_{\text{CE}} + \lambda \mathcal{L}_{\text{SD}},
\end{equation}
where $\lambda$ is a balancing factor that controls the relative importance of the GT supervision and the self-distillation loss. 

Notably, O2SP is only introduced during the training phase, meaning it does not incur any additional time or computational resource consumption during inference.
This self-distillation mechanism ensures that the feature representations of optical and SAR modalities are aligned in the feature space, facilitating more effective cross-modal learning and improving the overall performance of the CD model.

\section{Experiment} \label{B}
\subsection{Experimental Settings}
\paragraph{Dataset Description}
To validate the effectiveness of the proposed M$^2$CD framework, extensive experiments are conducted on the large-scale, publicly available optical-SAR multimodal CD dataset, CAU-Flood\cite{cau}. 
 The dataset consists of 18,302 image pairs, each with a size of 256×256 pixels, of which 15,231 pairs are used for training and 3,071 pairs for testing. From the training set, 3,807 pairs are randomly selected to form the validation set, ensuring an approximate ratio of 3:1:1 for training, validation, and testing, respectively.

\paragraph{Comparative Methods}
To demonstrate the superiority of the proposed M$^2$CD framework, we select 18 SOTA models for comparison. These models can be categorized into three groups based on their backbone structures and sizes: (1) CNN-based models, including FC-EF, FC-Siam-Conc, FC-Siam-Diff, STANet, IFN, SNUNet, BIT, TinyCD, HANet, CGNet, and the ResNet and ResNeSt versions of Changer; (2) Transformer-based models, including ChangeFormer and the MiT versions of Changer; and (3) foundation model-based methods, such as TTP. To further validate the generalizability of the proposed framework, we also design 5 variants of M$^2$CD based on ConvNeXt and MiT backbones. The backbones, detectors, parameter counts, and flops of these methods are summarized in Table \ref{table: comparison}.

\paragraph{Evaluation Metrics}  
To quantitatively evaluate the performance of the algorithms, we employ Precision (Prec), Recall (Rec), F1-score (F1), Intersection over Union (IoU), and Overall Accuracy (OA) metrics. Prec reflects the false positive rate, Rec reflects the false negative rate, and F1 balances these two metrics. IoU measures the overlap between predicted and GT regions, while OA provides an overall assessment of classification accuracy. 
We use $\text{TP}$, $\text{FP}$, $\text{TN}$, and $\text{FN}$ to denote true positives, false positives, true negatives, and false negatives, respectively. To comprehensively consider both change and non-change regions, we define $\text{Prec}_c$, $\text{Rec}_c$, $\text{Prec}_n$, and $\text{Rec}_n$ as the precision and recall for the change and non-change classes, respectively. 
OA and the mean versions of Prec, Rec, F1, and IoU can then be formulated as follows:
\begin{equation}
\begin{aligned}
&\text{OA} = \frac{\text{TP} + \text{TN}}{\text{TP} + \text{TN} + \text{FP} + \text{FN}},
\\
&\text{mPrec} = \frac{1}{2} \left( \frac{\text{TP}}{\text{TP} + \text{FP}} + \frac{\text{TN}}{\text{TN} + \text{FN}} \right),
\\
&\text{mRec} = \frac{1}{2} \left( \frac{\text{TP}}{\text{TP} + \text{FN}} + \frac{\text{TN}}{\text{TN} + \text{FP}} \right),
\\
&\text{mF1} = \frac{1}{2} \left( \frac{2 \times \text{Prec}_c \times \text{Rec}_c}{\text{Prec}_c + \text{Rec}_c} + \frac{2 \times \text{Prec}_n \times \text{Rec}_n}{\text{Prec}_n + \text{Rec}_n} \right),
\\
&\text{mIoU} = \frac{1}{2} \left( \frac{\text{TP}}{\text{TP} + \text{FP} + \text{FN}} + \frac{\text{TN}}{\text{TN} + \text{FN} + \text{FP}} \right).
\end{aligned}
\end{equation}

\paragraph{Implementation Details}  
To ensure fairness, we adopt the following identical settings for all methods: All experiments are conducted on a single NVIDIA 3090 GPU using the PyTorch-based OpenCD\cite{opencd} framework for both training and testing. 
During training, we employ data augmentation techniques, including RandomRotate, RandomFlip, and PhotoMetricDistortion. 
The batch size is set to 8 image pairs, and the patch size is fixed at 256×256, corresponding to the full image size. 
The AdamW optimizer is utilized with hyperparameters $\beta_1 = 0.9$ and $\beta_2 = 0.99$. 
All models are trained for 200k iterations, with validation performed every 1k iterations. 
The checkpoint achieving the best mIoU on the validation set is selected for testing. During testing, Test-Time Augmentation (TTA) is applied to enhance prediction robustness.

For the implementation of M$^2$CD, both the ConvNeXt and MiT backbones consist of four blocks. We integrate a MoE layer after each block, configuring the number of experts to 4 and selecting the top 2 experts as the output. 
The self-distillation weight $\lambda$ is empirically set to $1 \times 10^{-4}$.

\subsection{Experimental Results}
Table \ref{table: comparison} presents the numerical results of different methods. First, it is evident that our M$^2$CD with the MiT-b1 backbone achieves the best performance across all metrics except mRec. 
Benefiting from the large number of parameters in SAM, TTP achieves the second-best results. 
However, with less than 1/10 of its parameters and 1/25 of its FLOPs, our method outperforms TTP by 0.15\%, 0.36\%, and 0.57\% in OA, mF1, and mIoU, respectively. 
This demonstrates the superiority and efficiency of our M$^2$CD framework in multimodal CD tasks. 
Another interesting observation is that CNN-based models consistently underperform compared to Transformer-based models when dealing with optical-SAR modalities, including our M$^2$CD variants with ConvNeXt backbones. Nevertheless, our three ConvNeXt-based M$^2$CD models still outperform all other CNN-based methods, being the only ones to achieve an mIoU above 80\%, although with the highest parameter counts, which leaves room for future optimization.

\begin{table*}[!t]
\renewcommand{\arraystretch}{1.25}
\setlength{\tabcolsep}{10pt}
\caption{Comparison of Different Methods \label{table: comparison}}
\centering
\begin{tabular}{cccccccccc}
\hline
Method                        & Backbone    & Detector  & Param (M) & Flops (G) & OA             & mF1        & mPrec          & mRec        & mIoU           \\ \hline
FC-EF\cite{fcsn}                         & -           & FCN       & 1.353     & 3.244     & 89.73          & 77.23          & 79.26          & 75.59          & 66.04          \\ \hline
FC-Siam-Conc\cite{fcsn}                  & -           & FCN       & 1.548     & 4.989     & 91.59          & 79.59          & 86.24          & 75.59          & 68.95          \\ \hline
FC-Siam-Diff\cite{fcsn}                  & -           & FCN       & 1.352     & 4.385     & 86.04          & 46.30          & 83.52          & 50.02          & 43.05          \\ \hline
STANet\cite{stanet}                        & ResNet-18   & STAHead   & 12.764    & 17.578    & 76.58          & 53.82          & 53.67          & 54.07          & 43.90          \\ \hline
IFN\cite{ifn}                           & VGG-16      & -         & 35.995    & 78.982    & 93.66          & 86.42          & 87.57          & 85.37          & 77.44          \\ \hline
SNUNet\cite{snunet}                        & -           & FCN       & 3.012     & 11.73     & 94.60          & 88.05          & 90.72          & 85.83          & 79.77          \\ \hline
BIT\cite{bit}                           & ResNet-18   & BITHead   & 2.990     & 8.749     & 92.24          & 81.04          & 88.46          & 76.67          & 70.69          \\ \hline
\multirow{2}{*}{ChangeFormer\cite{changeformer}} & MiT-b0      & SegFormer & 3.847     & 2.455     & 95.09          & 89.70          & 89.99          & 89.42          & 82.16          \\
                              & MiT-b1      & SegFormer & 13.941    & 5.825     & 95.29          & 90.12          & 90.49          & 89.75          & 82.79          \\ \hline
TinyCD\cite{tinycd}                        & -           & -         & 0.285     & 1.448     & 94.12          & 86.83          & 90.02          & 84.29          & 78.06          \\ \hline
HANet\cite{hanet}                         & -           & FCN       & 3.028     & 20.822    & 93.78          & 86.13          & 89.03          & 83.79          & 77.09          \\ \hline
\multirow{5}{*}{Changer\cite{changer}}      & MiT-b0      & FDAF      & 3.457     & 1.741     & 95.53          & 90.63          & 90.93          & 90.34          & 83.57          \\
                              & MiT-b1      & FDAF      & 13.355    & 5.046     & 95.65          & 90.95          & 90.96          & 90.94          & 84.06          \\
                              & ResNet-18   & FDAF      & 11.391    & 5.955     & 91.51          & 83.24          & 81.68          & 85.08          & 73.05          \\
                              & ResNeSt-50  & FDAF      & 26.693    & 16.813    & 92.28          & 85.18          & 82.83          & 88.19          & 75.58          \\
                              & ResNeSt-101 & FDAF      & 47.485    & 29.469    & 89.91          & 76.04          & 80.80          & 73.00          & 64.91          \\ \hline
CGNet\cite{cgnet}                         & VGG-16      & -         & 38.989    & 87.55     & 94.25          & 87.08          & 90.49          & 84.40          & 78.41          \\ \hline
TTP\cite{ttp}                           & SAM       & SegFormer & 352.458   & 292.171   & {\ul 96.04}    & {\ul 91.60}    & {\ul 92.48}    & 90.78          & {\ul 85.09}    \\ \hline
\multirow{5}{*}{M$^2$CD}      & ConvNeXt-t  & SegFormer & 157.224   & 22.182    & 94.92          & 89.09          & 90.46          & 87.85          & 81.26          \\
                              & ConvNeXt-s  & SegFormer & 158.281   & 39.020    & 94.86          & 89.45          & 88.94          & 89.99          & 81.77          \\
                              & ConvNeXt-b  & SegFormer & 280.240   & 65.531    & 94.63          & 89.09          & 88.19          & 90.06          & 81.22          \\
                              & MiT-b0      & SegFormer & 6.989     & 6.372     & 95.68          & 91.14          & 90.59          & \textbf{91.72} & 84.35          \\
                              & MiT-b1      & SegFormer & 26.481    & 11.638    & \textbf{96.19} & \textbf{91.96} & \textbf{92.60} & {\ul 91.35}    & \textbf{85.66} \\ \hline
\end{tabular}
\end{table*}

\begin{figure*}[!t]
    \centering
    \includegraphics[width=7in]{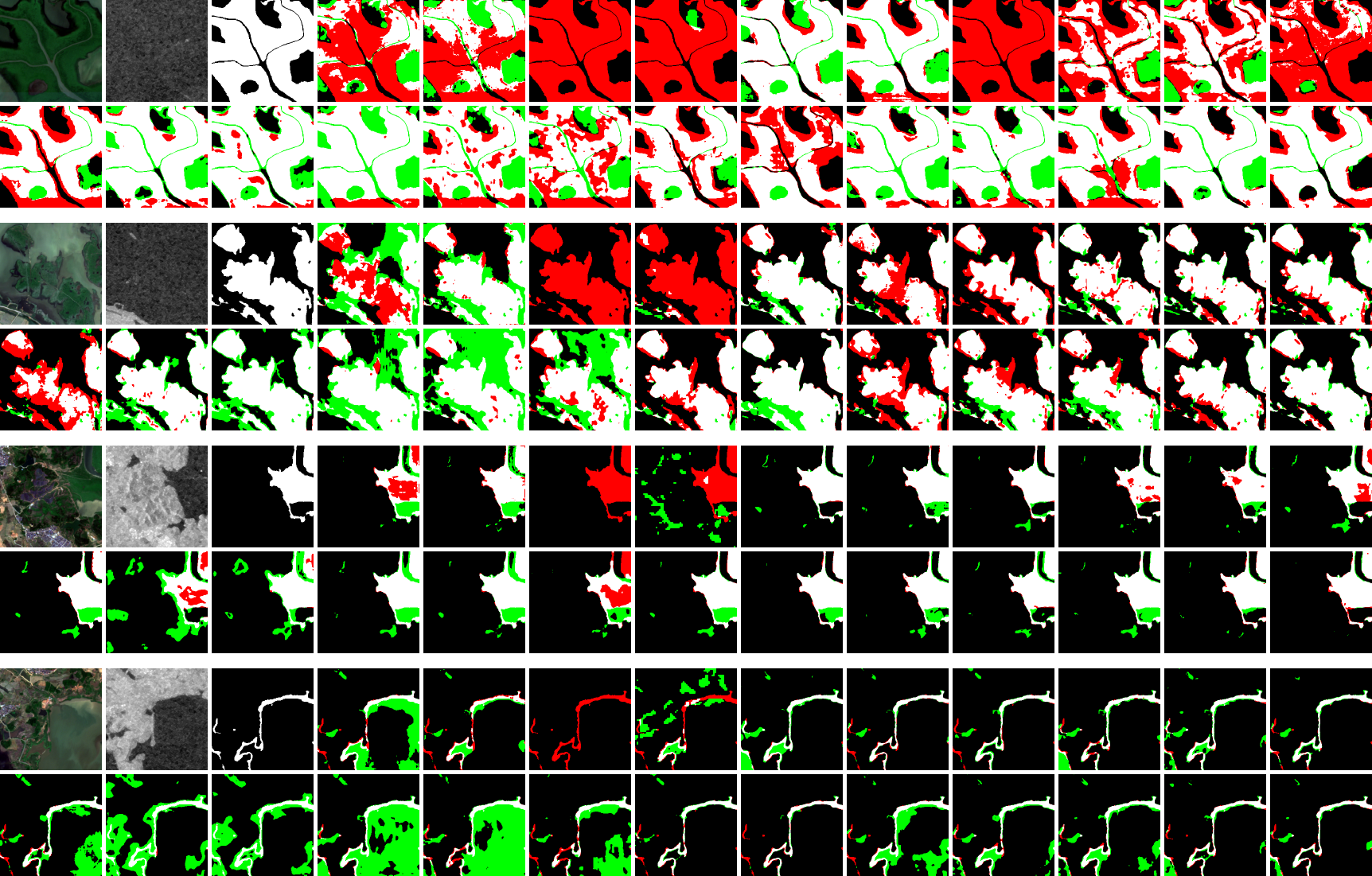}
    \caption{Visual comparison of different methods. For each set of images, from top-left to bottom-right: pre-event image, post-event image, ground truth, FC-EF, FC-Siam-Conc, FC-Siam-Diff, STANet, IFN, SNUNet, BIT, ChangeFormer-MiT-b0, ChangeFormer-MiT-b1, TinyCD, HANet, Changer-MiT-b0, Changer-MiT-b1, Changer-r18, Changer-s50, Changer-s101, CGNet, TTP, M$^2$CD-ConvNeXt-t, M$^2$CD-ConvNeXt-s, M$^2$CD-ConvNeXt-b, M$^2$CD-MiT-b0, and M$^2$CD-MiT-b1. Red indicates missed detections, and green indicates false alarms.}
    \label{fig:visual}
\vspace{-0.5cm}
\end{figure*}

\begin{table}[t]
\caption{Ablation on MoE and O2SP \label{table: ablation moe and O2SP}}
\centering
\begin{tabular}{ccccccc}
\hline
Backbone                & MoE                        & O2SP                       & mF1        & mPrec          & mRec        & mIoU           \\ \hline
\multirow{4}{*}{MiT-b0} & -                          & -                          & 90.16          & 90.77          & 89.59          & 82.86          \\
                        & \ding{51} & -                          & 90.75          & {\ul 90.92}    & {\ul 90.59}    & 83.76          \\
                        & -                          & \ding{51} & {\ul 90.88}    & \textbf{91.44} & 90.35          & {\ul 83.96}    \\
                        & \ding{51} & \ding{51} & \textbf{91.14} & 90.59          & \textbf{91.72} & \textbf{84.35} \\ \hline
\multirow{4}{*}{MiT-b1} & -                          & -                          & 90.20          & 90.48          & 89.92          & 82.91          \\
                        & \ding{51} & -                          & {\ul 91.84}    & {\ul 92.30}    & {\ul 91.39}    & 85.46          \\
                        & -                          & \ding{51} & 90.59          & 88.68          & \textbf{92.83} & 83.47          \\
                        & \ding{51} & \ding{51} & \textbf{91.96} & \textbf{92.60} & 91.35          & \textbf{85.66} \\ \hline
\end{tabular}
\vspace{-0.3cm}
\end{table}

Fig. \ref{fig:visual} visually demonstrates the performance of these methods when confronted with changes of varying scales caused by flooding. 
In the first set of images, we observe that many methods exhibit significant miss detections when dealing with large-scale changes, while M$^2$CD-MiT-b0 achieves the lowest missed detection rate and a lower false alarm rate compared to most methods. 
In the second set, M$^2$CD-MiT-b1 shows slightly higher missed detections than TTP but effectively reduces false alarms. 
SNUNet and M$^2$CD-ConvNeXt-s deliver the best performance in the third set of images.
The last set of images challenge the sensitivity of these algorithms to small-scale changes, where all methods exhibit varying degrees of false alarms and missed detections. 
M$^2$CD-MiT-b1 and TTP demonstrate superior connectivity in detecting narrow change regions, highlighting their effectiveness in identifying changes caused by water expansion or contraction.

\subsection{Ablation Study}  
We first validate the effectiveness of the M$^2$CD backbone. Table \ref{table: ablation moe and O2SP} presents the ablation results of the MoE and O2SP modules on the MiT-b0 and MiT-b1 backbones. The experimental results demonstrate that removing MoE or O2SP leads to a decline in model performance. On the other hand, using only MoE or O2SP can also improve the performance of the original backbone model.  

We further validate the effectiveness of M$^2$CD with different detectors, as shown in Table \ref{table: ablation detector}. Here, ``t, s, b'' represents the tiny, small, and base versions of ConvNeXt, respectively. 
For each backbone, we evaluate three detectors: FCN, UPerNet, and SegFormer. 
The results indicate that M$^2$CD consistently enhances model performance regardless of the detector used. 
Notably, on the original ConvNeXt-t and ConvNeXt-s models, UPerNet initially underperforms compared to SegFormer. However, under the M$^2$CD framework, UPerNet achieves superior performance, surpassing SegFormer.

\begin{table}[!t]
\caption{Ablation on Different Detectors \label{table: ablation detector}}
\centering
\begin{tabular}{cccccc}
\hline
Backbone           & Detector                   & Strategy & OA             & mF1        & mIoU           \\ \hline
\multirow{6}{*}{t} & \multirow{2}{*}{FCN}       & -        & 91.94          & 81.79          & 71.46          \\
                   &                            & M$^2$CD  & \textbf{93.35} & \textbf{85.76} & \textbf{76.53} \\ \cline{2-6} 
                   & \multirow{2}{*}{UperNet}   & -        & 93.97          & 87.24          & 78.58          \\
                   &                            & M$^2$CD  & \textbf{94.84} & \textbf{89.29} & \textbf{81.54} \\ \cline{2-6} 
                   & \multirow{2}{*}{SegFormer} & -        & 94.54          & 88.46          & 80.33          \\
                   &                            & M$^2$CD  & \textbf{94.92} & \textbf{89.09} & \textbf{81.26} \\ \hline
\multirow{6}{*}{s} & \multirow{2}{*}{FCN}       & -        & 92.79          & 84.93          & 75.38          \\
                   &                            & M$^2$CD  & \textbf{93.27} & \textbf{85.40} & \textbf{76.07} \\ \cline{2-6} 
                   & \multirow{2}{*}{UperNet}   & -        & 93.92          & 86.79          & 77.96          \\
                   &                            & M$^2$CD  & \textbf{94.83} & \textbf{89.42} & \textbf{81.72} \\ \cline{2-6} 
                   & \multirow{2}{*}{SegFormer} & -        & 94.32          & 88.08          & 79.77          \\
                   &                            & M$^2$CD  & \textbf{94.86} & \textbf{89.45} & \textbf{81.77} \\ \hline
\multirow{6}{*}{b} & \multirow{2}{*}{FCN}       & -        & 92.53          & 83.55          & 73.64          \\
                   &                            & M$^2$CD  & \textbf{93.12} & \textbf{85.24} & \textbf{75.83} \\ \cline{2-6} 
                   & \multirow{2}{*}{UperNet}   & -        & 94.32          & 87.94          & 79.57          \\
                   &                            & M$^2$CD  & \textbf{94.86} & \textbf{89.38} & \textbf{81.66} \\ \cline{2-6} 
                   & \multirow{2}{*}{SegFormer} & -        & \textbf{94.74} & 88.72          & 80.72          \\
                   &                            & M$^2$CD  & 94.63          & \textbf{89.09} & \textbf{81.22} \\ \hline
\end{tabular}
\vspace{-0.5cm}
\end{table}

\section{Conclusion} \label{C}
In this letter, we introduce a novel and unified multimodal CD framework, M$^2$CD, which significantly enhances the model's ability to process optical-SAR data through the integration of the MoE and O2SP self-distillation mechanisms. 
The MiT-b1 variant of our framework achieves SOTA performance in both visual results and quantitative metrics on the CAU-Flood dataset, a benchmark for optical-SAR CD. 
In future work, we aim to incorporate more modalities and collect more cross-modal image pairs to further advance multimodal RS and CD research.

\bibliographystyle{IEEEtran}
\bibliography{refs}

\end{document}